\documentclass[a4paper,conference]{IEEEtran}
\IEEEoverridecommandlockouts
\usepackage{multirow}
\usepackage{cite}
\usepackage{amsmath,amssymb,amsfonts}
\usepackage{algorithmic}
\usepackage{graphicx}
\graphicspath{ {./figs/} }
\usepackage{subfigure}
\usepackage{textcomp}
\usepackage{xcolor}

\usepackage{geometry}
\usepackage{array}
\usepackage{tabularx}
\usepackage{hhline}
\usepackage{xspace}
\usepackage{upgreek}
\usepackage{makecell}

\def\BibTeX{{\rm B\kern-.05em{\sc i\kern-.025em b}\kern-.08em
    T\kern-.1667em\lower.7ex\hbox{E}\kern-.125emX}}

\newcounter{itemlistc}

\begin{document}

\title{Physics-Enforced Modeling for Insertion Loss of Transmission Lines by Deep Neural Networks }

\author{\IEEEauthorblockN{Liang Chen$^1$ and Lesley Tan$^2$}
\IEEEauthorblockA{$^1$Department of Electrical and Computer Engineering, University of California, Riverside, CA 92521 USA\\
$^2$Phillips Academy, Andover, MA 01810 USA\\
liangch@ucr.edu, ltan22@andover.edu}
}

\maketitle

\begin{abstract}
In this paper, we investigate data-driven parameterized modeling of insertion loss for transmission lines with respect to design parameters. We first show that direct application of neural networks can lead to non-physics models with negative insertion loss. To mitigate this problem, we propose two deep learning solutions. One solution is to add a regulation term, which represents the passive condition, to the final loss function to enforce the negative quantity of insertion loss. In the second method, a third-order polynomial expression is defined first, which ensures positiveness, to approximate the insertion loss, then DeepONet neural network structure, which was proposed recently for function and system modeling, was employed to model the coefficients of polynomials. The resulting neural network is applied to predict the coefficients of the polynomial expression. The experimental results on an open-sourced SI/PI database of a PCB design show that both methods can ensure the positiveness for the insertion loss. Furthermore, both methods can achieve similar prediction results, while the polynomial-based DeepONet method is faster than DeepONet based method in training time.

\end{abstract}

\begin{IEEEkeywords}
deep neural networks (DNN), insertion loss, physics-enforced DNN (PDNN), polynomial-based DeepONet (PDeepONet)
\end{IEEEkeywords}

\section{Introduction}

Machine learning (ML), especially deep learning,  is gaining greater attention due to its success in various cognitive applications. Recently, machine learning approaches have been applied for electromagnetic simulation and modeling, which has shown great promise~\cite{Morten:IA'21}. Deep neural networks can be trained to build accurate models for nonlinear electromagnetic responses with different design parameters based on the simulated and experimental data.
However, such data-driven models based on deep neural networks or other machine learning methods may suffer the violation of some important physical properties, as those models may lack physical insights in the prediction of electromagnetic parameters. For instance, trained model based on the given data may fail to consider physics constraints, such as initial, boundary, and passive conditions. Therefore, it is important to address these issues for learning-based electromagnetic modeling to improve the physical prediction ability~\cite{Raissi:JCP'19}.

Recently, artificial neural networks (ANN) have been used to predict the S-parameter of high-speed interconnects based on an open SI/PI-Database~\cite{Morten:IA'21,Scharff:EPEPS'20}. However, the data-driven neural networks~\cite{Scharff:EPEPS'20} can lead to negative insertion loss, which violates the passive characteristics that were demonstrated in this paper. Insertion loss is a positive number that represents signal loss from the input power to the output power~\cite{Pozar2012}. Therefore, positive quantity is a physics constraint of insertion loss for the passive high-speed interconnects. 

Based on this observation, in this paper, we try to mitigate this problem by exploring advanced deep learning strategies. We propose two methods to enforce a positive quantity of insertion loss for ML-based parameterized transmission line models. In the first method, we propose to add a regularization item, which represents physics constraints of insertion loss, into the existing loss function to enforce positive quantity. This method is similar to the idea of physics-informed neural network concept (PINNs)~\cite{Raissi:JCP'19}. But here we just enforce the one specific physic law or constraint. In the second method, a third-order polynomial expression is applied to approximate the insertion loss first. The polynomial expression is designed so that it can guarantee the positiveness constraints on the insertion loss. Then, the neural network is applied to predict the coefficients of the polynomial expression instead, which shares a similar structure to recently proposed DeepONet structure for general function or dynamic system approximation~\cite{Lu:NMI'21}. The experimental results on an open-sourced SI/PI database of a PCB design show that both methods can ensure the positiveness for the insertion loss. Furthermore, the  physics-enforced DNN method and polynomial-based DeepONet method can achieve similar accuracy in both training and inferences, while the polynomial-based DeepONet method is faster than DeepONet-based method in training time.  

\section{Background and Motivation}
\label{sec:PF}
\begin{figure}[!ht]
    \centering
    \subfigure[]{\includegraphics[width=0.9\linewidth]{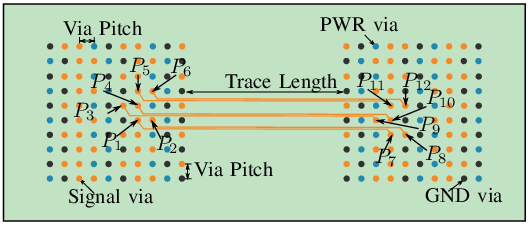}\label{fig:TopView}}
    \subfigure[]{\includegraphics[width=0.49\linewidth]{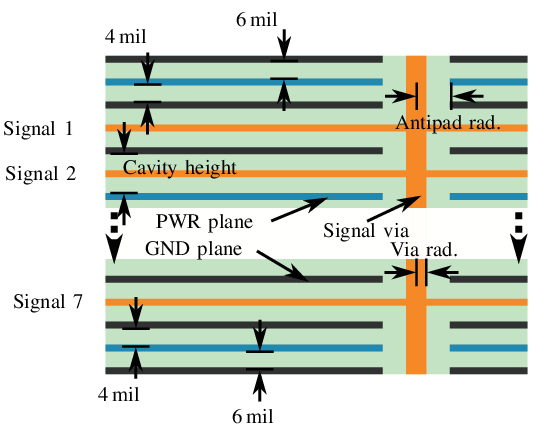}\label{fig:SideView}}
    \subfigure[]{\includegraphics[width=0.49\linewidth]{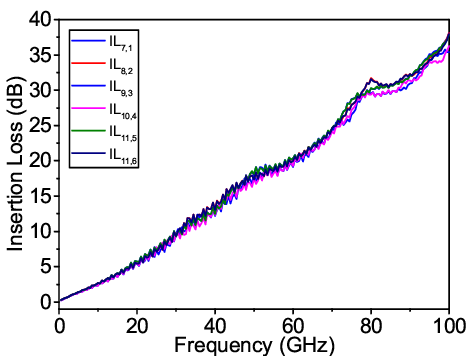}\label{fig:IL}}
    \caption{(a) Top view of the first signal layer on the PCB with a link between 2 via-arrays. (b) Side view. (c) The insertion loss for six interconnects with specific parameters.~\cite{Morten:IA'21}}
    \label{fig:Gemotry}
\end{figure}

It is well known that the insertion loss ($IL$) is defined as~\cite{Pozar2012}
\begin{equation}
    IL = -20\log_{10} |S_{21}| \text{ dB}
\end{equation}
where $S_{21}$ represents the transmission S-parameter of high-speed interconnects from port 1 to port 2. We perform our study on an open SI/PI dataset, which has transmission lines on 11 cavity PCB with two 10x10 via-arrays~\cite{Morten:IA'21}. Fig.~\ref{fig:TopView} shows the top view of the first signal layer on the PCB. There are six striplines with 12 ports, which provide a high-speed link between two 10x10 via-arrays. The side view of the PCB is shown in Fig.~\ref{fig:SideView}. The PCB consists of dielectric, vias, power planes, ground planes and signal layers. A cavity is formed between two adjacent power and ground planes. Fig.~\ref{fig:IL} shows the insertion loss of the six interconnects on the PCB with specific design parameters.

\begin{figure}[!ht]
    \centering
    \includegraphics[width=0.65\linewidth]{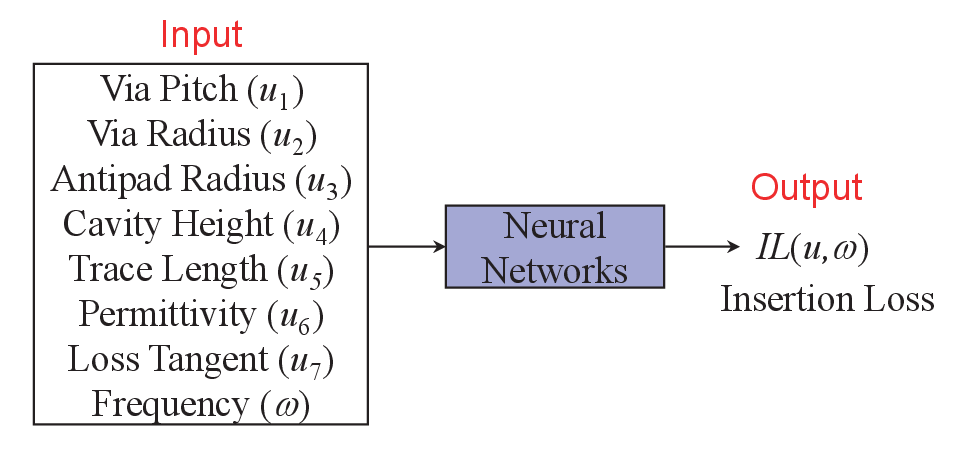}
    \caption{Fully connected neural networks for data-driven parameterized modeling of insertion loss of transmission lines with respect to design parameters.}
    \label{fig:FNN}
\end{figure}

We then try to develop the parameterized models of the insertion loss with respect to some design parameters. One straightforward solution is to apply fully connected neural networks, as shown in Fig.~\ref{fig:FNN}. For instance, we can model the insertion loss $IL_{71}$ of the interconnect from port 1 to port 7. The inputs and outputs of the neural networks are shown in Fig.~\ref{fig:FNN}.

\begin{figure}[!ht]
    \centering
    \subfigure[]{\includegraphics[width=0.49\linewidth]{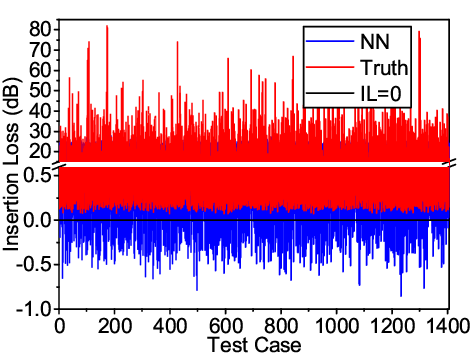}\label{fig:NN1}}
    \subfigure[]{\includegraphics[width=0.49\linewidth]{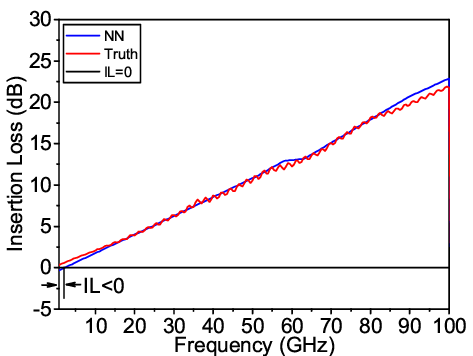}\label{fig:NN2}}
    \caption{(a) Predictions of insertion loss on test set by using NN, which show negative values (b)Insertion loss on one sample by using NN for different frequencies}
    \label{fig:NN}
\end{figure}

The dataset with 7030 samples is randomly split into a training set with 5624 samples (80\%) and test set with 1406 samples (20\%). To train the neural networks easily, we rescale all input and output features to [-1, 1] using the min-max normalization method. The predictions on the test set are shown in Fig.~\ref{fig:NN}. The mean-square-error (MSE) is 0.0145.
Fig.~\ref{fig:NN1} indicates that there are lots of samples which have negative insertion loss predicted by neural networks (NN). However, the results from ground truth are greater than 0. Therefore, NN fails to consider the positive quantity of insertion loss. The violation usually occurs at low frequency, as shown in Fig.~\ref{fig:NN2}. This issue should be addressed since it violates the passive characteristics.

\section{Physics-enforced Modeling of Insertion Loss}
To enforce positiveness of insertion loss, we propose to use two ML-based methods on the open SI/PI-Database. 
\subsection{Physics-enforced DNN}
In the physics informed neural networks, the initial conditions, boundary conditions and governing equations are considered by adding several loss functions. Based on the idea, an additional loss function representing the positive quantity of insertion loss is added to the total loss function of neural networks, which is expressed as
\begin{equation}
    Loss = Loss_{m}+\lambda\cdot Loss_{p}
\end{equation}
where $Loss_{m}$ is the mean-squared-error (MSE), which is represented by
\begin{equation}
   Loss_{m}=\frac{1}{N}\sum_{i=1}^N(IL^P_i-IL^T_i)^2
\end{equation}
where $IL^P$ and $IL^T$ are the prediction and ground truth of the insertion loss, respectively. $N$ is the number of sample points. $\lambda$ is a penalty coefficient. $Loss_{p}$ is the physics constraints of insertion loss, which is given by
\begin{equation}
   Loss_{p}=\max(0, -IL^P)
\end{equation}
which represents that insertion loss is not less than 0. The auxiliary information can guide neural networks to train the model so that it satisfies the physics constraints.

\subsection{Polynomial-based DeepONet}
\begin{figure}[!ht]
  \centering
  \includegraphics[width=0.8\columnwidth]{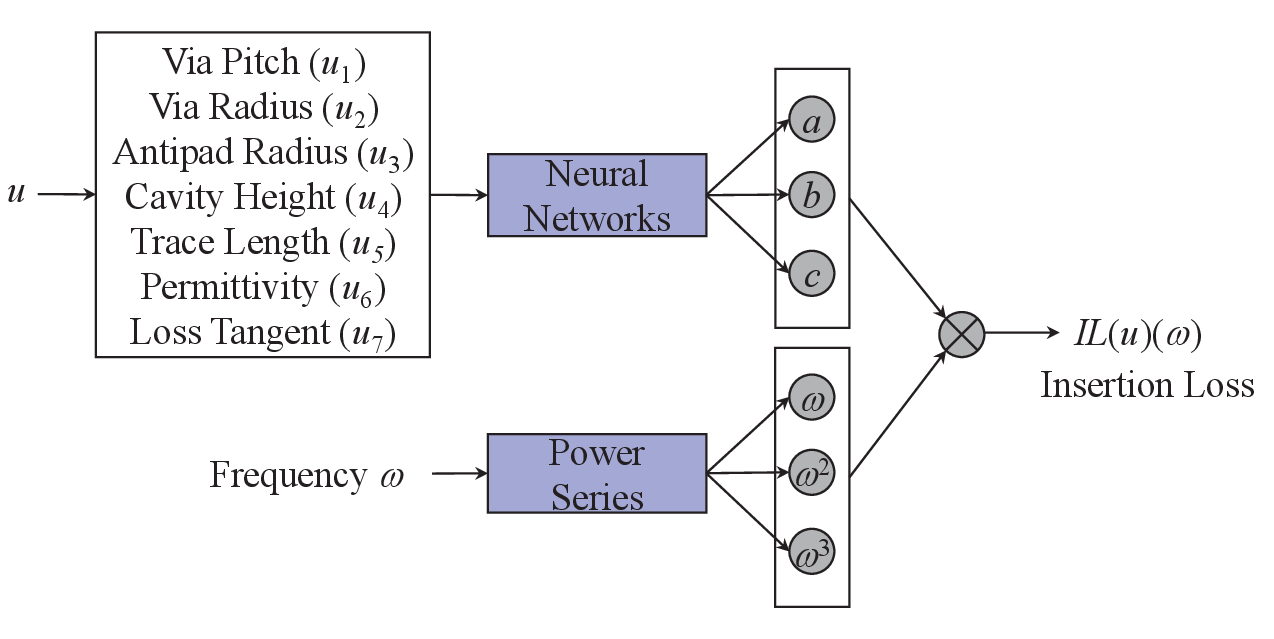}
  \caption{The new architecture of neural networks based on polynomial approximation.}
  \label{fig:framework}
\end{figure}

The neural networks can be capable of fitting any functions and it can easily find several solutions of the problem. Some solutions may not meet the physics constraints. To avoid this situation, we can first define the form of the solution. In this method, we use a third-order polynomial expression to approximate the insertion loss~\cite{Koledintseva:TEMC'14}, which is written as
\begin{equation}
    IL(\omega) = a\cdot\omega+b\cdot\omega^2+c\cdot\omega^3
\end{equation}
where $a$, $b$, and $c$ are the coefficients, and $\omega$ is the frequency.
Then, we apply the neural networks to find the coefficients of the expression.
The new architecture of neural networks is shown in Fig.~\ref{fig:framework}.
The input and output of neural networks are dimensional parameters and coefficients, respectively. Power series module generates frequency to the power of 1, 2, and 3. After that, the insertion loss is obtained by the multiplication of the frequency power series and corresponding coefficients. The new networks in Fig.~\ref{fig:framework} can be represented by
\begin{equation}
    \bigg|IL(u)(\omega)-\sum_{i=1}^3 \text{NN}_i(u) \omega^i\bigg|<\epsilon
\end{equation}
where NN$_i(u)$ is the $i$th variable of the neural networks' outputs. Note in our implementation, we first use the fitting method in MATLAB to obtain the coefficients $a$, $b$, and $c$, which becomes functions of the input design parameters. Then, we train the neural networks for the three functions $a$, $b$, and $c$.


\section{Experimental Results}
\begin{figure}[!ht]
    \centering
    \subfigure[]{\includegraphics[width=0.49\linewidth]{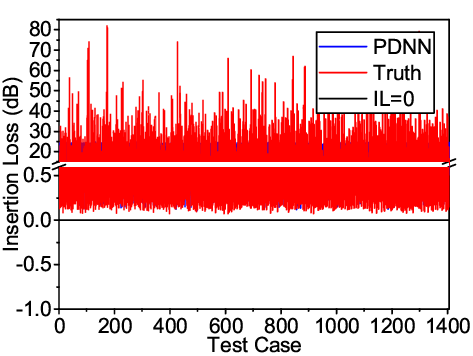}\label{fig:PINN}}
    \subfigure[]{\includegraphics[width=0.49\linewidth]{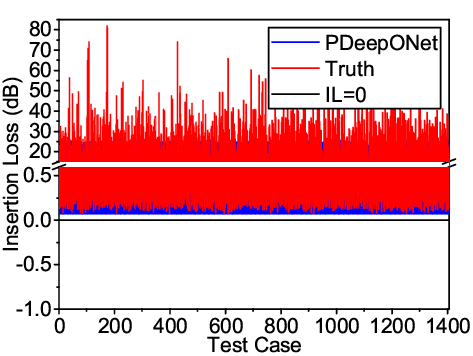}\label{fig:EPA}}
    \caption{Predictions of insertion loss on test set by using (a) PDNN and (b) PDeepONet.}
\end{figure}

In this section, we demonstrate the two proposed methods to enforce the positiveness of insertion loss on the same test set, which is illustrated in Section~\ref{sec:PF}. Fig.~\ref{fig:PINN} shows the results predicted by Physics-enforced DNN, called {\it PDNN}. The minimum values of predictions is almost the same as ground truth. There is no negative value in the predictions and thus this method satisfies the physics constraints of insertion loss. Fig.~\ref{fig:EPA} shows that the predicted insertion loss by the polynomial-based DeepONet ({\it PDeepONet}) method for different samples. As we can see, no negative values are predicted. This is because the defined polynomial expression gives the constraint on insertion loss. In summary, the two ML-based methods ensure the positiveness of insertion loss on test set.
\begin{figure}[!ht]
  \centering
  \includegraphics[width=0.7\columnwidth]{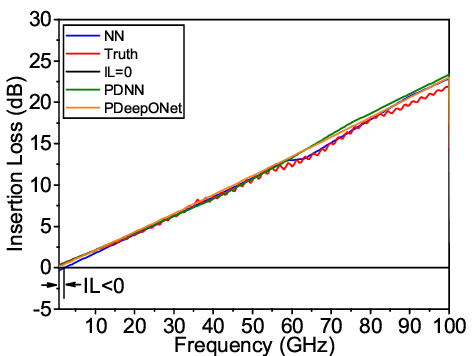}
  \caption{Comparison of insertion loss on one sample with different methods.}
  \label{fig:comparison}
\end{figure}

To further compare the two methods, we test them on one sample in test cases. The results are shown in Fig.~\ref{fig:comparison}. The negative insertion loss usually occurs at low frequency for the {\it NN} method. It can be observed that the proposed two methods can fix the problem. 

Although the two methods can predict positive insertion loss, their performances are different, which are illustrated in Table~\ref{table:comparison}. The training and test mean squared errors (MSEs) of {\it PDNN} are 0.015 and 0.0145, respectively, which are the same as those of {\it NN}. The errors of {\it PDeepONet} are slightly larger than those of {\it NN} and {\it PDNN}. The inference of {\it PDeepONet} costs the maximum time even though its training time is less than those of {\it NN} and {\it PDNN}. Because {\it PDeepONet} needs to consider the computation time in MATLAB. It can be seen from Table~\ref{table:comparison} that time for {\it PDeepONet} consists of two parts which are time in neural networks and time in MATLAB. The {\it PDNN} spends more training time than {\it NN} since it takes extra time to satisfy the physics constraints. One caveat we want to point out is that different implementations of the two methods may change the time differences  in both training and inferences and the observed trend also need to be further investigated for more datasets in the future. 
\begin{table}[h!]
\centering
\caption{Accuracy and speed comparison}
\begin{tabular}{|c|c|c|c|c|}
\hline
\multirow{2}{*}{}                                             & \multicolumn{2}{c|}{Training set}                                              & \multicolumn{2}{c|}{Test set}                                             \\ \cline{2-5} 
                                                              & MSE    & Time (s)                                                       & MSE    & Time (s)                                                     \\ \hline
NN                                                            & 0.0150 & 603.71                                                         & 0.0145 & 0.021                                                        \\ \hline
\begin{tabular}[c]{@{}c@{}}PDNN\end{tabular} & 0.0150 & 1851.32                                                        & 0.0145 & 0.021                                                        \\ \hline
PDeepONet                                                    & 0.0151     & \begin{tabular}[c]{@{}c@{}}46.05+379.50\\ =425.55\end{tabular} & 0.0146 & \begin{tabular}[c]{@{}c@{}}0.012+0.076\\ =0.088\end{tabular} \\ \hline
\end{tabular}
\label{table:comparison}
\end{table}

\section{Conclusion}
\label{sec:con}
In this paper, we have proposed two methods to enforce the positive quantity of insertion loss predicted by machine learning-based methods. The first one is physics-enforced DNN method. We add an additional loss function, which represents the passive condition, to the final loss function to enforce negative quantity of insertion loss. In the second method, a third-order polynomial expression is defined to approximate the insertion loss. Then, the neural network is applied to predict the coefficients of the polynomial expression. The experimental results on an open-sourced SI/PI database of transmission line design showed that both methods can ensure the positiveness for the insertion loss. Furthermore,  both methods can achieve similar prediction results, while the polynomial-based DeepONet method is faster in training time.

\bibliographystyle{IEEEtran}
\bibliography{./bib/refer.bib}

\end{document}